\documentclass[letter, 10pt, conference]{ieeeconf}      %
\IEEEoverridecommandlockouts                              %
\makeatletter
\newcommand*\titleheader[1]{\gdef\@titleheader{#1}}
\AtBeginDocument{%
  \let\st@red@title\@title
  \def\@title{%
    \bgroup\normalfont\large\centering\@titleheader\par\egroup
    \vskip1.5em\st@red@title}
}
\makeatother

\usepackage{amsmath} %
\usepackage{amssymb}  %
\usepackage[textsize=scriptsize]{todonotes}
\usepackage{multirow}
\usepackage{cite}
\usepackage[final]{changes}
\usepackage{textcomp}
\usepackage[font=footnotesize,skip=5pt]{caption}

\title{\LARGE \bf
	Wheelless Soft Robotic Snake Locomotion: Study on Sidewinding and Helical Rolling Gaits
}
\titleheader{\small{This paper has been accepted to 2023 IEEE-RAS International Conference on Soft Robotics (RoboSoft).}}
\author{Dimuthu~D. K.~Arachchige, Dulanjana M. Perera$^{1}$, Sanjaya Mallikarachchi, \\Iyad Kanj, Yue Chen$^{2}$, and Isuru~S.~Godage$^{3}$%
	\thanks{$\!\!\!\!\!\!$School of Computing, Jarvis College of Computing and Digital Media, DePaul University, Chicago, IL 60604, USA.\,
		\newline Corresponding author: {\tt\small DARACHCH@depaul.edu}\,
		\newline
        $^{1}$Department of Multidisciplinary Engineering, Texas A\&M University, College Station, TX 77843, USA.\, \newline
		$^{2}$Department of Biomedical Engineering, Georgia Institute of Technology, Atlanta, GA 30332, USA.\,\newline
		$^{3}$Department of Engineering Technology \& Industrial Distribution and J. Mike Walker \textquotesingle66 Department of Mechanical Engineering, Texas A\&M University, College Station, TX 77843, USA.
        \vspace{1mm}		
        \newline
        This work is supported in part by the National Science Foundation (NSF) Grants IIS–2008797, CMMI–2048142, and CMMI–2132994.
	}
}

\begin{document}
	
	\maketitle
	\thispagestyle{empty}
	\pagestyle{empty}
	
	\begin{abstract}

Soft robotic snakes (SRSs) have a unique combination of continuous and compliant properties that allow them to imitate the complex movements of biological snakes. Despite the previous attempts to develop SRSs, many have been limited to planar movements or use wheels to achieve locomotion, which restricts their ability to imitate the full range of biological snake movements.
We propose a new design for the SRSs that is wheelless and powered by pneumatics, relying solely on spatial bending to achieve its movements. We derive a kinematic model of the proposed SRS and utilize it to achieve two snake locomotion trajectories, namely sidewinding and helical rolling. These movements are experimentally evaluated under different gait parameters on our SRS prototype.
The results demonstrate that the SRS can successfully mimic the proposed spatial locomotion trajectories. This is a significant improvement over the previous designs, which were either limited to planar movements or relied on wheels for locomotion. The ability of the SRS to effectively mimic the complex movements of biological snakes opens up new possibilities for its use in various applications.

	\end{abstract}

	\section{Introduction} \label{sec:Introduction}

	Snakes are among the few reptile species that do not require limbs to locomote in various environments including, marshes, deserts, and dense vegetation. Within the powerful musculature that generates movements, snakes have a skeletal structure that protects internal organs and facilitates smooth bending to produce unique locomotion patterns \cite{gray1946mechanism,zhen2015modeling}. 
	Further, the small cross-section-to-length ratio of snakes facilitates passing through confined and narrow spaces. Thus, robotic snakes, both rigid and soft, inspired by their biological counterparts are ideally suited for applications such as search and rescue operations and inspection tasks\cite{yamauchi2022development}. 	
	Soft robotic snakes (SRS) can generate smooth body deformation (bending) and are more adaptable to their surroundings than rigid robotic snakes due to their inherent compliance and continuous structures, making them the best candidate to emulate natural snake locomotion. 
	
	\begin{figure}[t] 
		\centering
		\includegraphics[width=1\linewidth,height=0.55\linewidth]{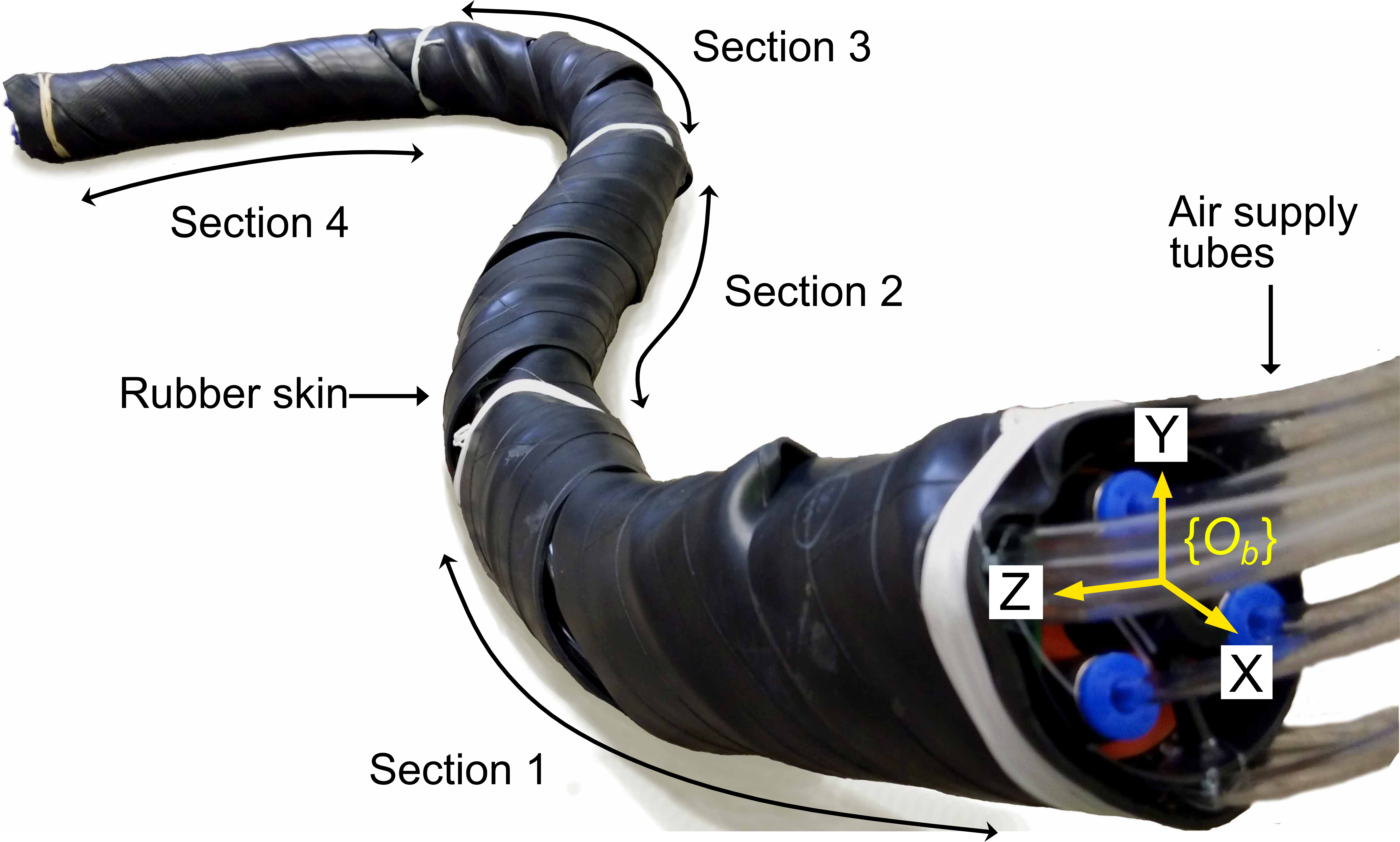}
		\caption{Four-section soft robotic snake prototype.}
		\label{fig:Fig1_IntroductionImage} 
	\end{figure}
	
	Snakes use friction anisotropy between snakeskin and the moving surface to generate forward propulsion necessary for planar locomotion \cite{hu2009mechanics}.	
	But, typical SRS skin cannot generate anisotropic frictional forces, unless the snakeskin is improved by other means. 
	Thus, some SRS prototypes, such as those reported in \cite{onal2013autonomous,luo2014theoretical,luo2015slithering,luo2017toward} utilize axially-mounted passive wheels to generate anisotropic frictional forces and achieve locomotion. The SRSs appeared in \cite{branyan2018soft,branyan2020snake,qi2020novel,qi2022bioinspired} showed wheelless planar locomotion achieved through improved snakeskin with anisotropic frictional properties. 
	There are SRSs that exhibit spatial deformation and achieve lateral undulation, sidewinding, and step-climbing gaits \cite{luo2018orisnake,wan2022design}. However, they require passive wheels attached to their bodies. Wheels are cumbersome and can hinder locomotion. Additionally, wheels prevent achieving other types of snake locomotion gaits. Hence, wheeled SRSs are not suitable to study organic snake locomotion. 
	
	Note, no SRS that shows spatial rolling gaits exists. In our previous work \cite{arachchige2021soft}, we proposed a pneumatically actuated SRS that was capable of spatial bending but was limited to achieving planar gaits including undulation and rolling. The SRS only had three bending sections and thus lacked adequate degrees of freedom (DoF) to lift the body off the ground and simultaneously maintain sufficient ground contact forces to generate a motion. Further, the bending sections do not contain a backbone and are thus subjected to length change during bending which can cause undesirable reaction forces that counteracted the forward progression.
	
	Snakes tend to use spatial gaits when friction anisotropy does not exist (i.e., in deserts). For example, snakes use sidewinding to spatially move the body while minimizing concentrated ground contacts (similar to articulated limbs of octopi). 
	In this work, we introduce spatial bendability as an alternative to overcome limitations associated with the anisotropic frictional forces and wheels of SRSs. Similar to snakes' spatial gaits, the idea is to maintain skin-ground contact at a minimum level. %
	We propose an SRS prototype with four sections (Fig. \ref{fig:Fig1_IntroductionImage}) to circumvent the limitations of the previous SRS design \cite{arachchige2021soft}. Notably, the new SRS is inextensible as the bending sections have integrated backbones.
	The presence of a backbone makes the SRS design more bioinspired because similar to snakes it has a skeletal structure to support the locomotion. Extending \cite{arachchige2021soft}, we, in this work, i) design and fabricate a novel 4-section SRS, ii) present a complete kinematic model of the SRS, iii) generate locomotion trajectories for sidewinding and helical rolling motions, and iv) experimentally validate them on the SRS prototype. To the best of the authors' knowledge, this is the first demonstration of wheelless spatial locomotion and helical rolling gaits for SRSs. 
	
	\section{Soft Robotic Snake Kinematic Modeling}%
	
	\subsection{{Soft Robotic Snake Prototype}\label{subsec:SRS_Prototype}}
	
	\begin{figure}[tb] 
		\centering
		\includegraphics[width=1\linewidth,height=0.68\linewidth]{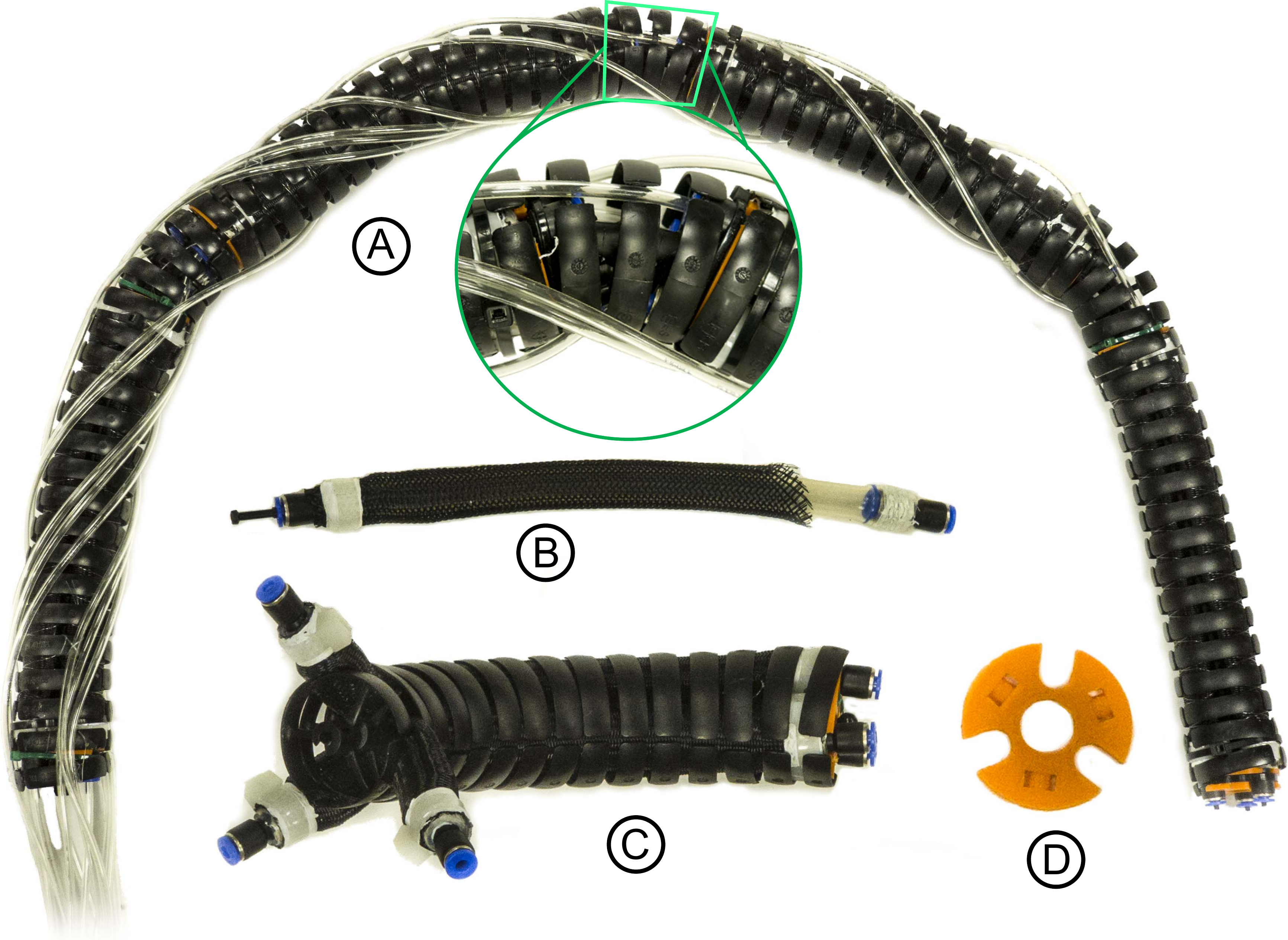}
		\caption{(A) SRS prototype: serially arranged four soft bending robot sections without its rubber skin showing spherically routed pressure supply tubes on the backbone outer shell. The enlarged image shows the backbone offset between adjacent sections. (B) A PMA--braided sleeve on the right side has been removed to show the Silicone tube sealed to a push-to-connect pneumatic union fitting. (C) One section with its PMA arrangement inside the backbone. (D) PMA mounting plates of each section.}
		\label{fig:Fig2_PartDesign} 
	\end{figure}

	We adopt three, actuated DoF, hybrid soft robotic design approach proposed in \cite{arachchige2022hybrid} to fabricate the 4-section SRS prototype shown in Fig. \ref{fig:Fig1_IntroductionImage}. %
	An SRS section (of length $240~mm$ and radius $20~mm$) is made of a flexible backbone and three McKibben-type extending-mode pneumatic muscle actuators (PMAs)~\cite{godage2012locomotion}.
	A PMA is fabricated using a flexible Silicone tube dressed with a braided sleeve which then is sealed by push-to-connect pneumatic union fittings at either end (Fig. \ref{fig:Fig2_PartDesign}B). 
	A cable carrier (Triflex R-TRL40, Igus), which is a rigid kinematic chain, is chosen as the flexible backbone. 
	PMAs are tri-symmetrically placed inside the backbone around its neutral axis as shown in Fig. \ref{fig:Fig2_PartDesign}C. 
	We mount either side of the PMA ends to the backbone using two 3D-printed end-caps (Fig. \ref{fig:Fig2_PartDesign}D). 
	The thickness of the end caps is set to $5~mm$ to retain the flexibility of the backbone to the fullest extent. 
	An SRS section bends in a constant curvature arc due to its inextensible backbone \cite{arachchige2022hybrid}. 
	
	We serially connect adjacent sections with a backbone length offset $\left(50~mm\right)$, creating a hollow skeletal area between each section (Refer to the enlarged image in Fig. \ref{fig:Fig2_PartDesign}A). Accordingly, the complete SRS has three identical length offsets between four sections along its body. These offsets -- made as extensions of the backbone length itself -- facilitate connecting pressure supply tubes to PMAs of each section while preserving the continuum nature of the complete SRS assembly. 
	We 
	wrap pressure supply tubes within the SRS body without obstructing its bendability as shown in Fig. \ref{fig:Fig2_PartDesign}A. 
	To obtain a uniform snakeskin (hence uniform friction), a thin rubber sleeve is wrapped around the SRS body (Fig. \ref{fig:Fig1_IntroductionImage}). It covers routed pressure supply tubes and eliminates their adversarial effects during locomotion. The SRS has 12-DoF (3-DoF in each section) in total relative to its base. Without pressure supply tubes, the total weight and the length of the SRS prototype are $0.94~kg$ and $1.11~m$, respectively.
	
	The hybrid design approach adopted here enables a higher stiffness control range with adequate structural integrity necessary for the SRS spatial locomotion \cite{arachchige2021novel, amaya2021evaluation,arachchige2022hybrid,xiao2023kinematics,azizkhani2023dynamic,godage2019center,gilbert2019validation}. 
	It should be noted, the backbone-integrated bending units (i.e., SRS sections) are heavier than the ones without a backbone. 
	Hence, it is required to generate higher torques to overcome friction and weight during spatial bending. Thus, the presence of a backbone enables us to achieve bending at higher stiffnesses without significant torsion in SRS sections. 
	
	\subsection{Complete Robot System Model}\label{subsec:SRS-System-Model}

	\begin{figure}[tb] 
		\centering
		\includegraphics[width=1\linewidth,height=0.75\linewidth]{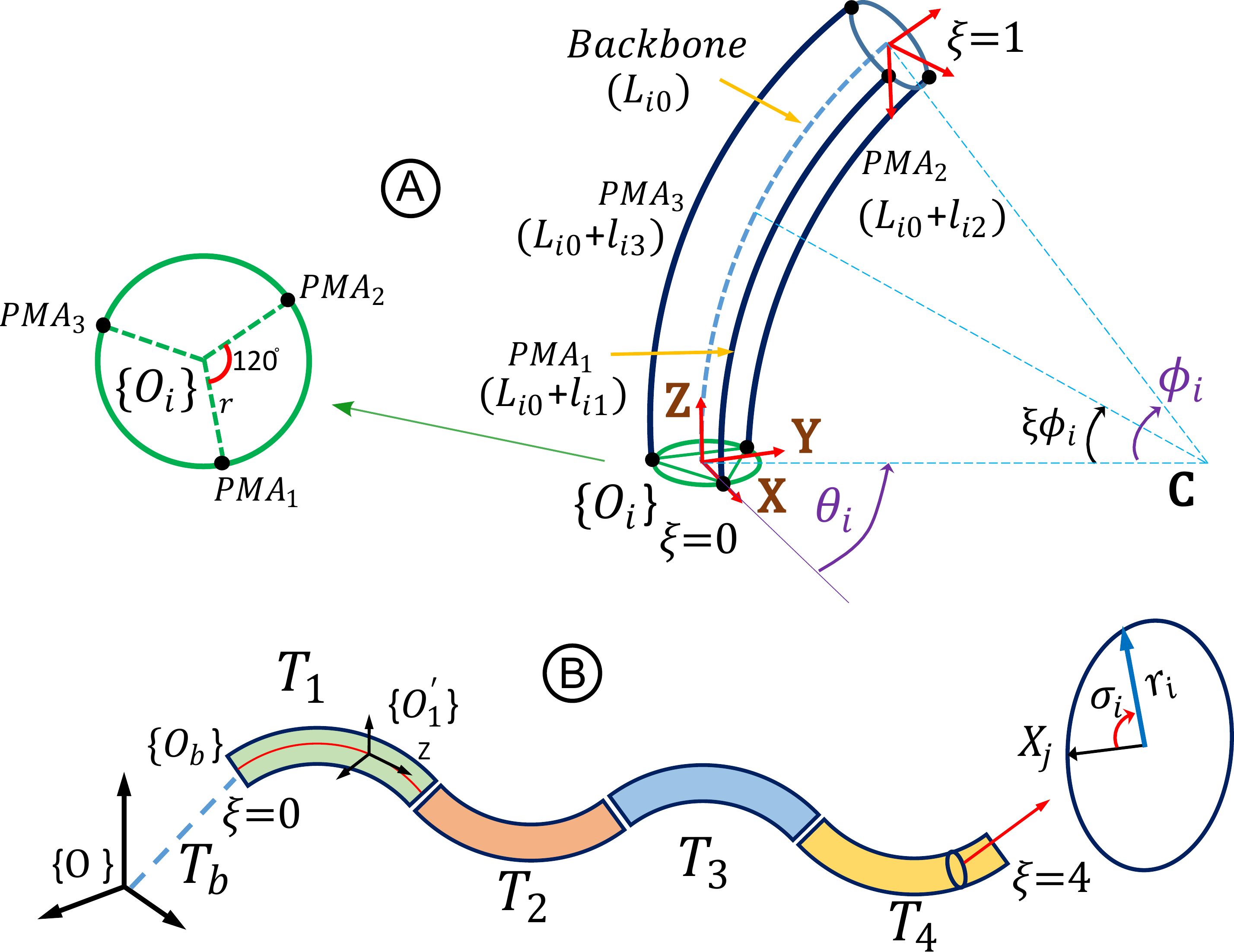}
		\caption{ The SRS schematics. (A) An SRS section showing its PMA arrangement. (B) 4-section SRS showing the global, robot, and local coordinates frames and kinematic parameters.}
		\label{fig:Fig3_SchematicDiagram} 
	\end{figure}

    \begin{figure}[tb] 
		\centering
		\includegraphics[width=1\linewidth]{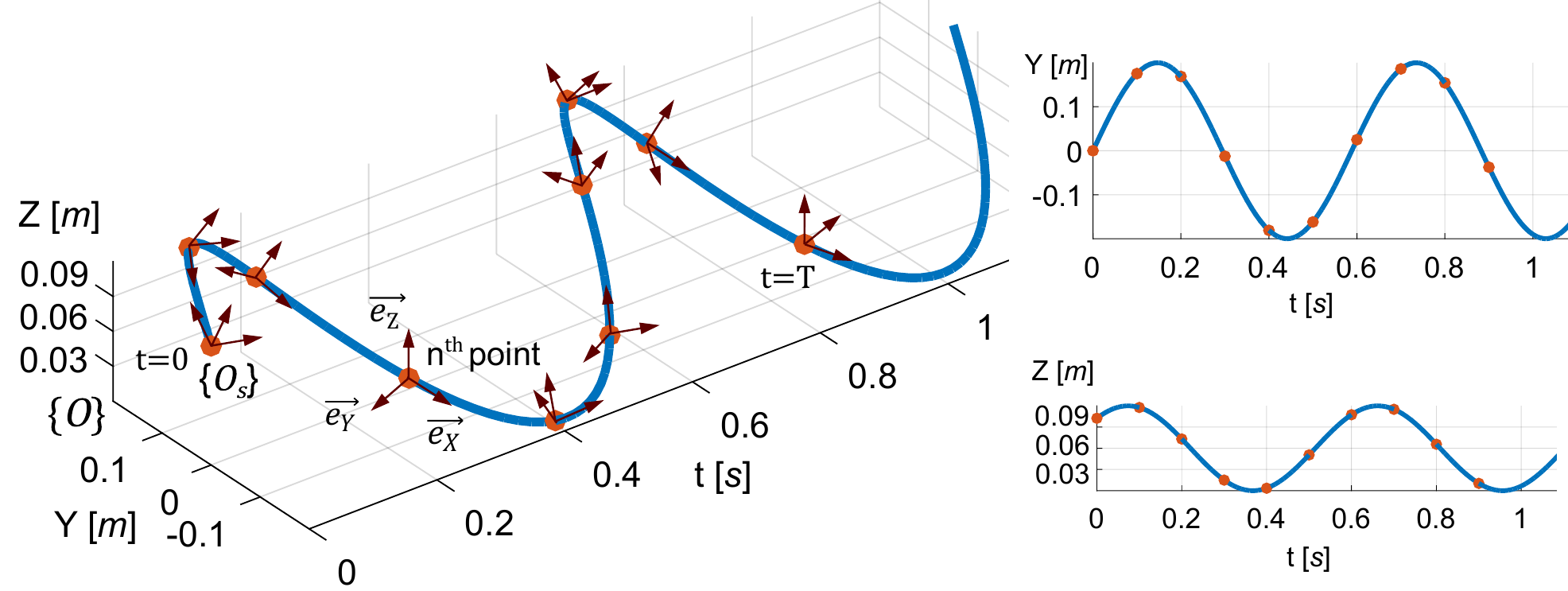}
		\caption{Periodic discretization of the sidewinding curve relative to the global frame, \{$O$\}.}
		\label{fig:Fig4_SidewindingTrajGeneration} 
	\end{figure}
	
	In the proposed SRS, the kinematics of a single SRS section can be serially extended to obtain the kinematics of the complete SRS. 
	The kinematics of a single section has been in detail discussed in \cite{deng2019near,arachchige2022hybrid,chen2021modal}. Herein, we extend their work to obtain the kinematics of the SRS skin.
	Considering any \textit{i}-th section $\left(i\in \{1,2,3,4\}\right)$ of the SRS, Fig. \ref{fig:Fig3_SchematicDiagram}A illustrates the kinematic diagram thereof. 
	The coordinate frame $\{O_{i}\}$ is placed at the center of the \textit{i}-th section and the $Z$ axis is aligned with the backbone (i.e., neutral axis).
	The length changes of PMAs, $l_{ij} \in \mathbb{R}$ $\left(j\in \{1,2,3\}\right)$, causes the section to bend in $\theta \in [-\pi,~\pi]$ bending direction and $\phi\in [0,~\pi]$ bending angle \cite{deng2019near,arachchige2022hybrid}.
	Therefore, the total length of a PMA can be expressed as $L_{ij}=L_{i0}+l_{ij}$, where $L_{i0}$ is the backbone length. 
	Additionally, due to constant curvature bending, a special kinematic constraint, $l_{i1} + l_{i2} + l_{i3} = 0$ must be satisfied \cite{deng2019near}.
	Accordingly, one out of three, kinematic DoF is redundant and we can express the jointspace variable for the \textit{i}-th section as $q_{i} = [l_{i1}, l_{i2}]^{T}$. 
	By adopting modal kinematics proposed in \cite{godage2015modal}, we can derive the homogeneous transformation matrix (HTM), $\mathbf{T_{i}}\in \mathbb{SE}\left(3\right)$ of any point on the surface (skin) of the \textit{i}-th section as
	\vspace{-0.5mm}
	\begin{align}
		\mathbf{T}_{i}\left(\boldsymbol{q}_{i},\xi_{i}\right) & =\left[\begin{array}{cc}
			\mathbf{R}_{i}\left(\boldsymbol{q}_{i},\xi_{i}\right) & \mathbf{p}_{i}\left(\boldsymbol{q}_{i},\xi_{i}\right)\\
			\mathit{\boldsymbol{0}} & 1
		\end{array}\right]\cdots\nonumber \\
		& \qquad\left[\begin{array}{cc}
			\mathbf{R}_{z}\left(\sigma_{i}\right) & \mathcal{\mathit{\boldsymbol{0}}}\\
			\mathit{\boldsymbol{0}} & 1
		\end{array}\right]\left[\begin{array}{cc}
			\boldsymbol{1} & \mathbf{p}_{x}\left(r_{i}\right)\\
			\mathit{\boldsymbol{0}} & 1
		\end{array}\right]\label{eq:ith_kin}
	\end{align}
	where $\mathbf{R}_{i}(\sigma_{i})\in\mathbb{SO} \left(3\right)$ and $\mathbf{p}_{i}\in\mathbb{R}^{3}$ denotes the rotation matrix and position vector of a point on the neutral axis. 
	
	A point on the skin of the \textit{i}-th section is obtained by translating HTM in \eqref{eq:ith_kin} by $\mathbf{R}_{z}\in\mathbb{SO} \left(3\right)$ and $\mathbf{p}_{x}(r_{i})\in\mathbb{R}^{3}$ where $\sigma_{i}\in [0,~2\pi]$ and $r_{i}\in\mathbb{R}$ are the angular offset $\left(Z^{+}\right)$ and the position offset $\left(X^{+}\right)$ of the surface point from the neutral axis. 
	$\xi_{i}\in\left[0,1\right]$ is a selection factor for points along the neutral axis such that 0 gives the base point and 1 gives the tip location.
	When the SRS moves, its base at $\{O_{b}\}$ floats on the global coordinate frame, $\{O\}$. By combining a floating-base coordinate frame, $\mathbf{T}_{b}$, we define the kinematic model of the complete SRS as
	\vspace{-0.5mm}
	\begin{align}
		\mathbf{T}\left(\boldsymbol{q}_{b},{\boldsymbol{q}_r},{\xi}\right) & =\mathbf{T}_{b}\left(\boldsymbol{q}_{b}\right)\prod_{i=1}^{4}\mathbf{T}_{i}\left({\boldsymbol{q}_{i}},\xi_{i}\right)\nonumber \\
		& =\left[\begin{array}{cc}
			\mathbf{R}\left(\boldsymbol{q}_{b},{\boldsymbol{q}_r},\xi\right) & \mathbf{p}\left(\boldsymbol{q}_{b},{\boldsymbol{q}_r},\xi\right)\\
			\mathit{\mathbf{0}} & 1
		\end{array}\right]\label{eq:complete_kin}
	\end{align}
	where $\boldsymbol{q}_{r}=[\boldsymbol{q}_{1},\boldsymbol{q}_{2},\boldsymbol{q}_{3},\boldsymbol{q}_{4}]\in \mathbb{R}^{12}$ is the vector that contains all joint variables and ${\xi}=\left[0,4\right]\in\mathbb{R}$ selects a point along the SRS (i.e., SRS base at $\xi=0$, and SRS tip  at $\xi=4$). $\boldsymbol{q}_{b}=\left[x_{b},y_{b},z_{b},\alpha,\beta,\gamma\right]\in\mathbb{R}^{6}$ defines translation and angular offsets of $\{O_{b}\}$ relative to $\{O\}$ (Fig. \ref{fig:Fig3_SchematicDiagram}).
	
	\section{Locomotion Trajectory Generation\label{sec:Trajectory-Generation}}
	
	\subsection{Methodology\label{subsec:TrajectoryGenerationProcedure}}
	
	We generate locomotion trajectories to move the robot. Snakes show periodic locomotion, and hence, their locomotion gaits can be modeled as cyclic mathematical curves. 
	First, we mathematically model the desired locomotion gait (i.e., trajectory) on the global coordinate frame 
	(i.e., robot taskspace). Next, one period of the identified trajectory curve (i.e., the mathematical curve) is discretized and the curve at discretized locations is projected onto the robot coordinate frame; this is referred to as trajectory sampling and coordinate transformation (from the global frame to the local frame). Finally, the inverse kinematic solution of \eqref{eq:complete_kin} is applied to convert the local taskspace (i.e., local curve points -- $x,y,z$) into jointspace variables required to actuate the robot. The above procedure is applied to obtain jointspace trajectories in Secs. \ref{subsec:Sidewinding-Trajectory} and \ref{subsec:3D-Rolling-Trajectory}.

	\begin{figure}[tb] 
		\centering
		\includegraphics[width=1\linewidth]{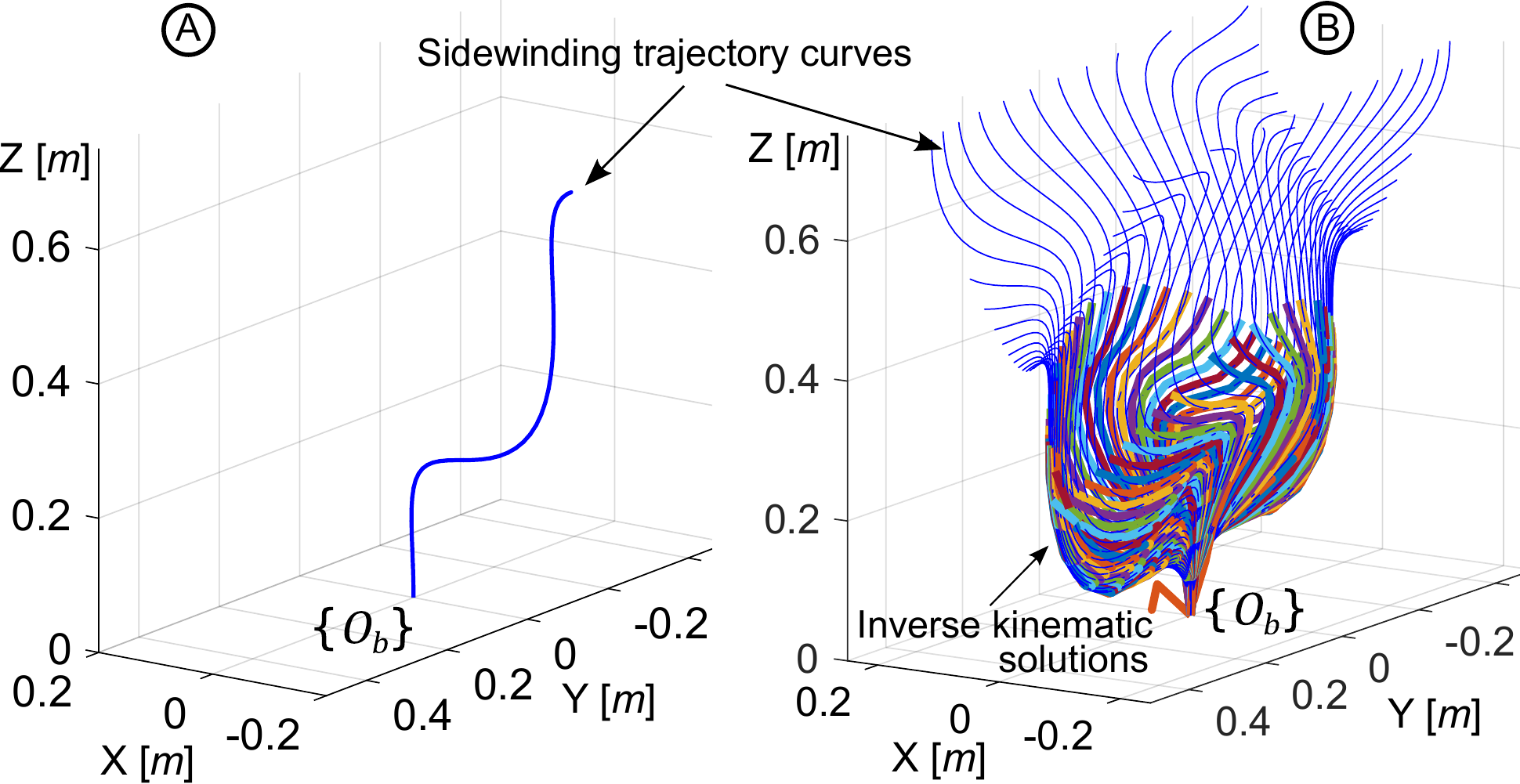}
		\caption{In sidewinding trajectory: (A) A discretized curve at $t=0$ projected onto the robot coordinate frame, $\{O_b\}$. (B) Generated trajectory curves after projecting all discretized curve points onto \{$O_b$\} and matched jointspace trajectories (i.e., inverse kinematic solutions).}
		\label{fig:Fig5_SidewindingMathematicalCurves} 
	\end{figure}

 \begin{figure}[tb] 
		\centering
		\includegraphics[width=1\linewidth]{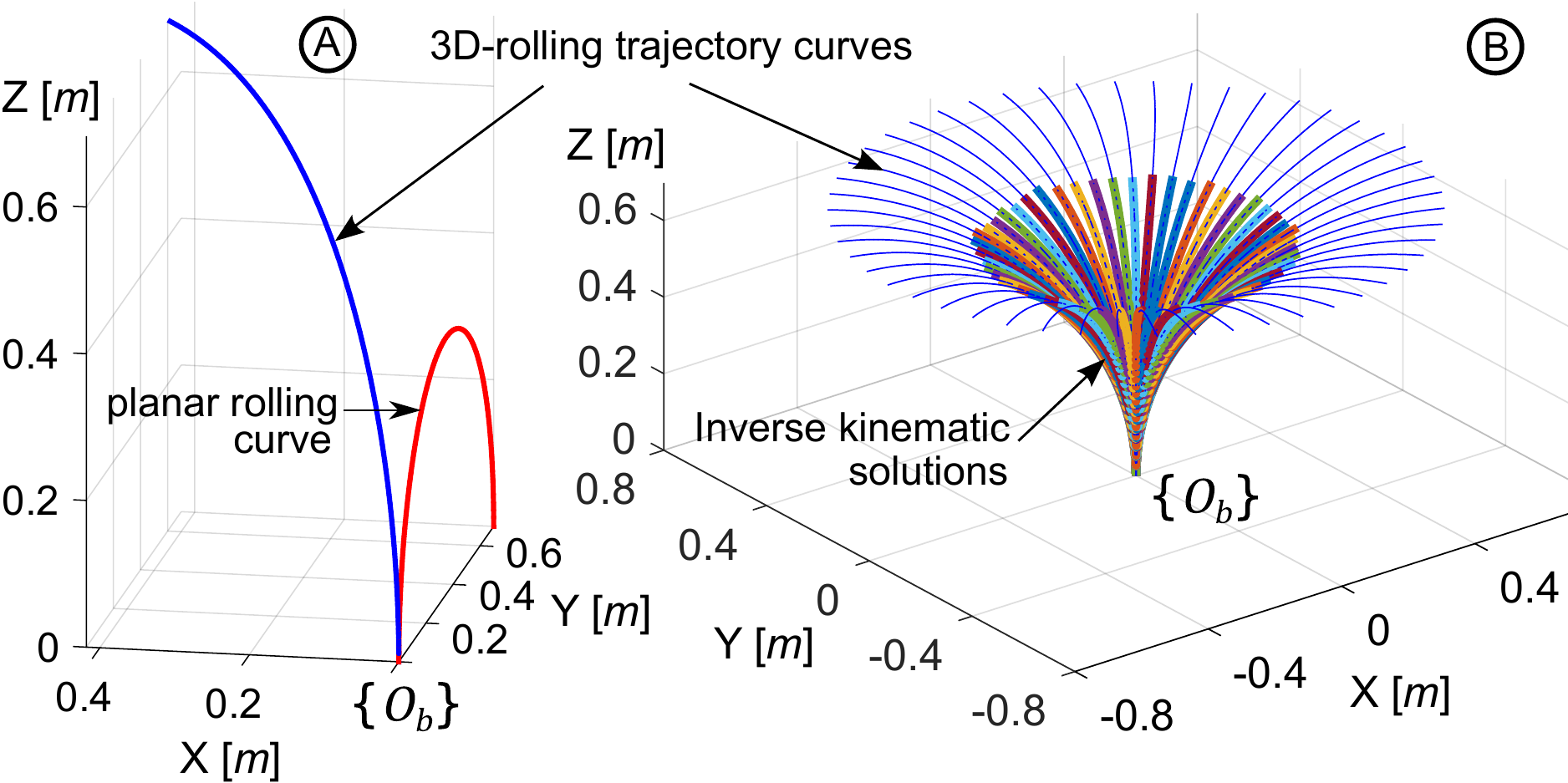}
		\caption{(A) A trajectory curve projected onto \{$O_b$\} in planar and helical rolling. (B) Projected total trajectory curves onto \{$O_b$\} and generated jointspace trajectories (inverse kinematic solutions) in helical rolling.}
		\label{fig:Fig6_3DRollingMathematicalCurve} 
	\end{figure}
 
	\subsection{Sidewinding Gait Trajectory\label{subsec:Sidewinding-Trajectory}}

	A time-series $\left(t\right)$ sidewinding curve is given by, 
	\vspace{-0.5mm}
	\begin{align}
		\begin{split}y\left(t\right) & =A_{y}\sin\left(2\pi f_{y}t\right)\\
			z\left(t\right) & =A_{z}\sin\left(2\pi f_{z}t+\beta\right)
		\end{split}
		& \label{eq:sidewindingcurve}
	\end{align}	
	where $A_y,\,A_z$, and $f_y,\,f_z$ define amplitudes and frequencies of $y$ and $z$ curve segments, respectively. Herein, $\beta$ denotes the phase shift between two curve segments. 
	
	Fig. \ref{fig:Fig4_SidewindingTrajGeneration} shows the progression of the sidewinding curve derived with
	$A_y=0.2~m,A_z=0.05~m,f_y=2~Hz,f_z=2~Hz,\beta=\frac{\pi}{3}~rad$, and period, $T$. We decided on those values based on the physical dimensions of the SRS prototype. In Fig. \ref{fig:Fig4_SidewindingTrajGeneration}, $\{O_s\}$ and $\{O\}$ define the curve's local coordinate frame and the global coordinate frame, respectively. First, we uniformly discretize the curve at marked locations such that those points designate the origin of the robot coordinate frame, $\{O_{b}\}$ at the discretized time instances. Next, we derive local coordinate frames at each point relative to $\{O\}$. At any $n^{th}$ point, the local coordinate frame can be defined as a tangential line to the curve (let it be local $X$) and a line normal to that (let it be local $Y$). If $\overrightarrow{e}_{X}$ and $\overrightarrow{e}_{Y}$ define unit vectors along the local $X$ and local $Y$, respectively, then $\overrightarrow{e}_{X} \times \overrightarrow{e}_{Y}$ gives the unit vector along the local $Z$. We encapsulate this orientation data with the discretized position data to derive the local HTM at each point relative to $\{O\}$ along the curve. Then, these local HTMs at each instance are utilized to project the sidewinding curve onto $\{O_b\}$. Fig. \ref{fig:Fig5_SidewindingMathematicalCurves}A shows the projection of the taskspace curve at the time instance, $t=0$ onto $\{O_b\}$. Similar to that, all subsequent taskspace curves at each discretized time instance within the trajectory period, $T$ are projected onto $\{O_b\}$ as depicted by thin blue lines in Fig. \ref{fig:Fig5_SidewindingMathematicalCurves}B.
	
	\begin{figure}[tb] 
		\centering
		\includegraphics[width=1\linewidth]{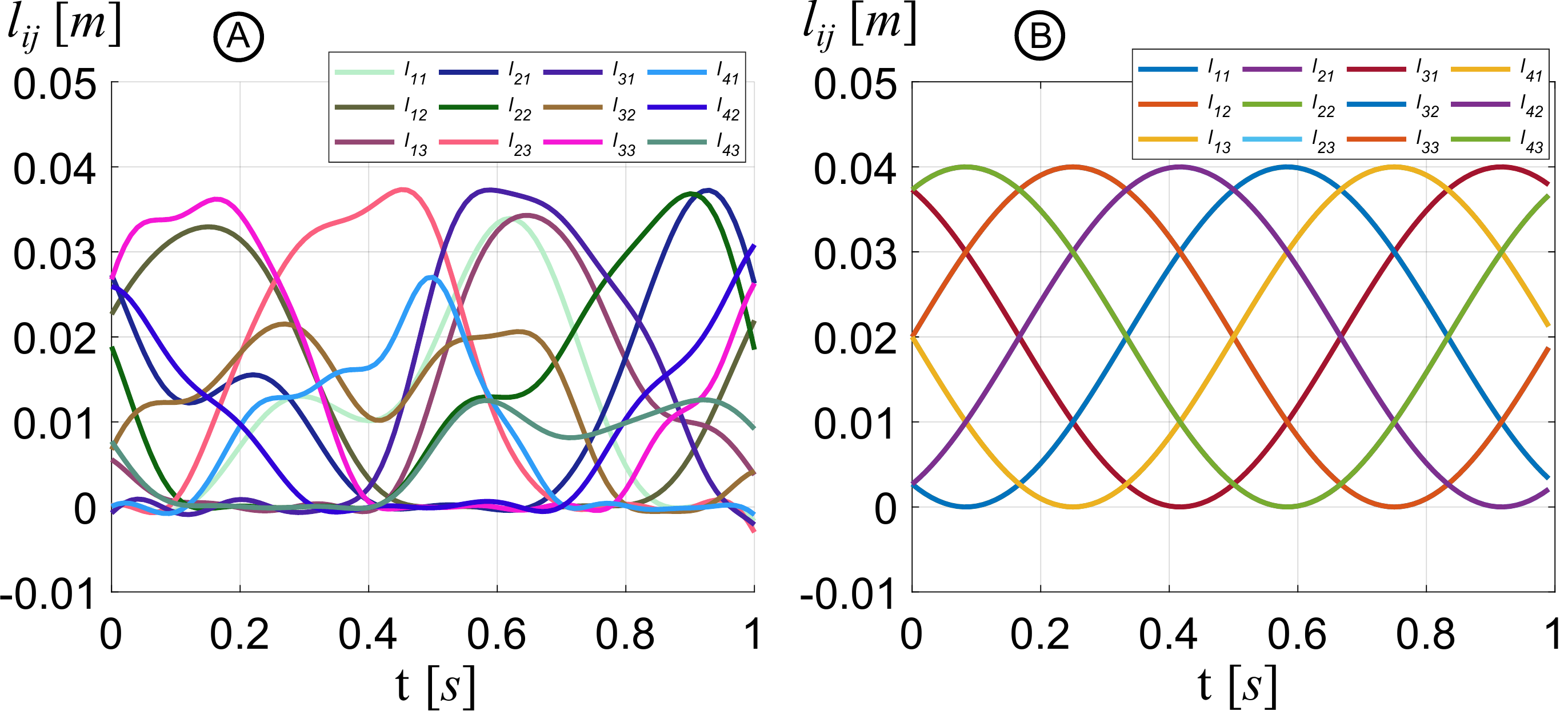}
		\caption{Optimization results, i.e., matched jointspace trajectories (length changes) for (A) Sidewinding and (B) Helical rolling gaits within a trajectory period, $T=1~s$. Note: In helical rolling, the length variables of section 1 and section 4 overlap each other due to the phase shift applied.} 
		\label{fig:Fig7_LengthTrajectories} 
	\end{figure}
	
	\begin{figure}[tb] 
		\centering
		\includegraphics[width=1\linewidth]{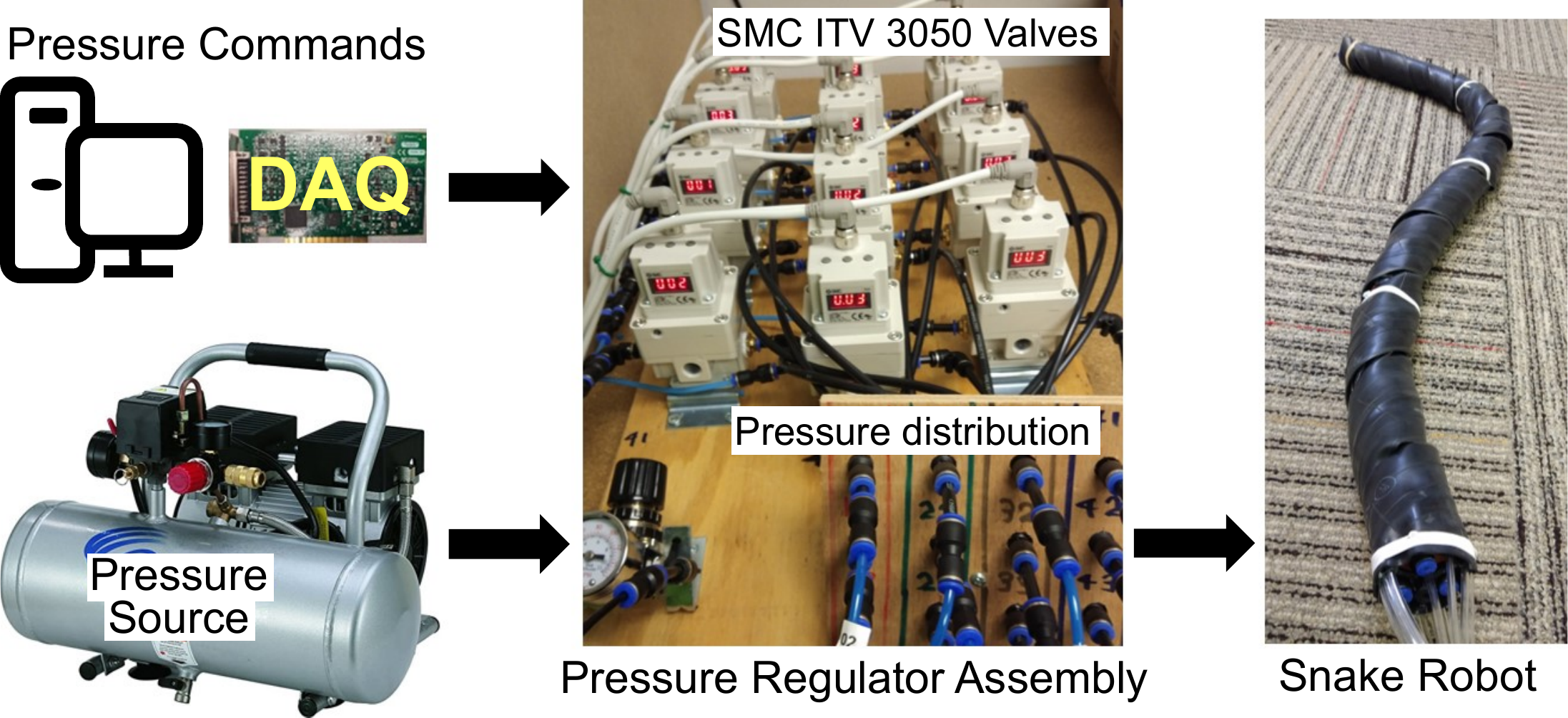}
		\caption{ SRS experimental setup.}
		\label{fig:Fig8_ExperimentalSetup} 
	\end{figure}
	
	Finally, we obtain the SRS joint variables by mapping the trajectory curves at $\{O_b\}$ (i.e., local taskspace curves) into the SRS jointspace using inverse kinematics~\cite{meng2022rrt,meng2021smooth}. However, it is impossible to obtain closed-form inverse kinematic solutions for multi-section continuum robots \cite{godage2015modal} such as the 4-section SRS proposed here. For that reason, we formulate the solution for the inverse kinematics as an optimization problem between the desired local trajectory curves and the SRS forward kinematics~\cite{godage2012path}. Therein, we apply the kinematic model in \eqref{eq:complete_kin} to obtain uniformly distributed points (61 points -- 15 per section) along the SRS neutral axis. Correspondingly, the cost function is formulated as
	\vspace{-1.5mm}
	\begin{align}
		f_{cost} &=\sum_{k=1}^{61}\left\Vert \mathbf{p}\left({0},{q},\xi_k\right)-f_{gait}\left(s\right)\right\Vert +\sum_{j\in\left\{ 1,2,3\right\}}^{i\in\left\{ 1,2,3,4\right\}} l_{ij}^{2}\label{eq:costfun} 
	\end{align}
	where, $\xi_k=\xi$ given in \eqref{eq:complete_kin} is discretized into 61 points along the SRS and $f_{gait}\left(s\right)$ is the trajectory curve (with $s=61$ discretized points) to which the SRS should fit. 
	
	The latter term of \eqref{eq:costfun} ensures the stability and the smoothness of the optimized solution. We implemented \eqref{eq:costfun} using MATLAB’s global constrained optimization routine. Fig. \ref{fig:Fig5_SidewindingMathematicalCurves}B shows the matched SRS shapes (thick multi-color lines) with the projected taskspace curves (thin blue lines) for the sidewinding gait. 
	
	\subsection{Helical Rolling Gait Trajectory\label{subsec:3D-Rolling-Trajectory}}
	
	Helical rolling (i.e., 3D\slash Spatial rolling) is an extension of the planar rolling presented in our previous work \cite{arachchige2021soft}. Here, first, we briefly discuss the adopted trajectory generation procedure of the planar rolling and later, extend it to the helical rolling. In \cite{arachchige2021soft}, we modeled the planar rolling as circular arc displacements on a plane. Therein, we defined a trajectory cycle as a curve rotation about the SRS's neutral axis (i.e., Z of $\{O_{b}\}$). Due to the robot's girth, the rotation results in a displacement on the X-Y plane. This arc displacement within a period was discretized into several circular arc shapes at uniformly distributed time instances (Fig. 3-b in \cite{arachchige2021soft}). Next, we projected those taskspace rolling curves at each time instance onto the robot coordinate frame, $\{O_b\}$. In the end, we applied the optimization method proposed in \eqref{eq:costfun} and obtained jointspace trajectories. Readers are referred to \cite{arachchige2021soft} for more details on deriving planar rolling trajectories. 
	
	\begin{figure*}[tb] 
		\centering
		\includegraphics[width=1\textwidth, height=0.275\textwidth]{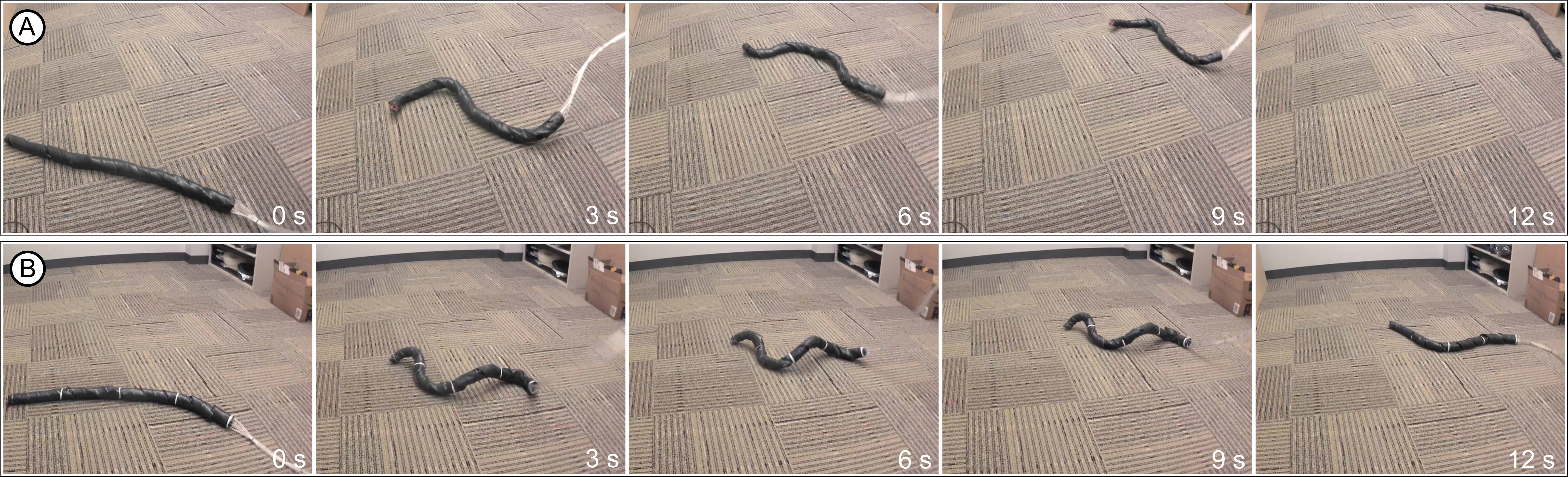}
		\caption{The SRS progression for (A) Sidewinding gait at $4~bar-1.00~Hz$, and (B) Helical rolling gait at $4~bar-0.50~Hz$.}
		\label{fig:Fig9_RollingSidewindingExperiments} 
	\end{figure*}
	
	We generate the helical rolling trajectories by applying the same procedure of the planar rolling. However, in this case, we actuate adjacent SRS sections with an angular phase shift, $\vartheta=\frac{\pi}{3}$. Accordingly, the cumulative phase shift of each adjacent section (sections 2, 3, and 4) becomes $\frac{\pi}{3},\frac{2\pi}{3},$ and $\pi$, respectively, relative to the first section. Because of that, the SRS spatially moves out of its current plane along the X, Y, and Z directions relative to its base at $\{O_b\}$, which does not otherwise occur in planar rolling as illustrated in Fig. \ref{fig:Fig6_3DRollingMathematicalCurve}A. Note that, no angular phase shift is applied to the first SRS section relative to its base. Due to applied phase shifts, the SRS is expected to maintain skin-ground contacts at three points. We mathematically derive the taskspace curve of the helical rolling on the global framework, $\{O\}$ by applying $\vartheta=\frac{\pi}{3}$ between adjacent sections of the planar rolling curve. 
	Similar to Sec. \ref{subsec:Sidewinding-Trajectory}, the derived curve is discretized and the curves at discretized locations are projected onto $\{O_b\}$  and matched for jointspace variables as shown in Fig. \ref{fig:Fig6_3DRollingMathematicalCurve}B. %
	Figs. \ref{fig:Fig7_LengthTrajectories}A and \ref{fig:Fig7_LengthTrajectories}B show optimization results, i.e., jointspace trajectories (or length changes) for sidewinding and helical rolling gaits during one cycle. Here, the SRS actuation rate (i.e., gait frequency) and the SRS bending can be controlled by adjusting the gait period and the jointspace amplitude, respectively.
	
	\section{Results and Discussion\label{sec:Results_and_Discussion}}
	
	\begin{figure}[tb] 
		\centering
		\includegraphics[width=1\linewidth]{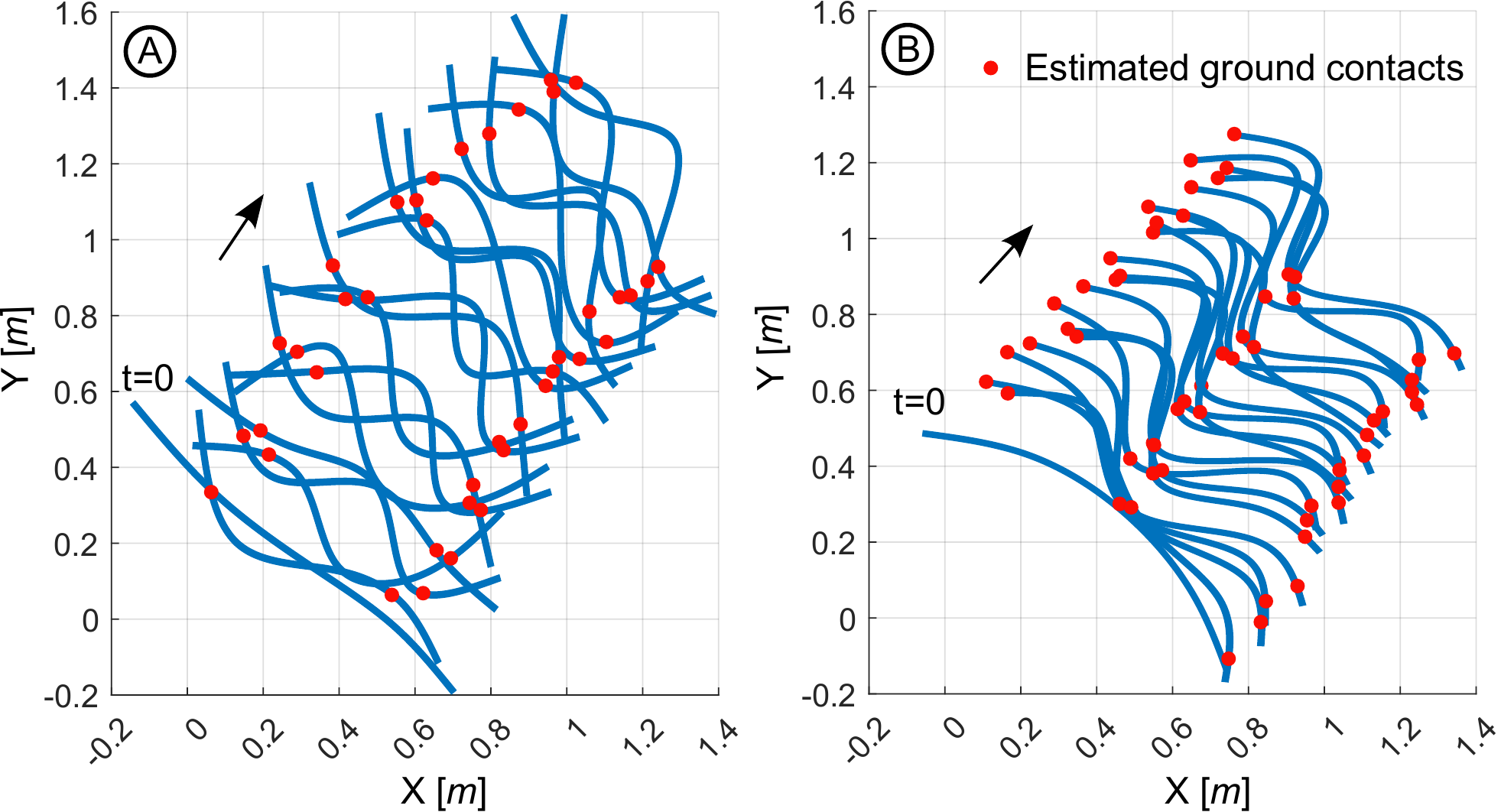}
		\caption{Locomotion tracking: (A) Sidewinding, (B) Helical rolling.}
		\label{fig:Fig10_LocomotionTracking} 
	\end{figure}
	
	\subsection{Experimental Setup \label{subsec:Experimental_Setup}}
	
	The experimental setup employed in validation studies is shown in Fig. \ref{fig:Fig8_ExperimentalSetup}. 
	The air compressor provides steady $8~bar$ pressure to 12 digital proportional pressure regulators (SMC ITV3050) that are then connected to individual PMAs (12 PMAs in 4 sections) of the SRS. The MATLAB Simulink Real-Time model computes the length changes and generates $0-10~V$ voltage signals via a data acquisition card (National Instruments PCI-6221) to control the pressure regulators in real-time at $20~Hz$. During each test, the SRS prototype was actuated for $12~s$ on a carpeted floor with uniform friction.
	Note that, the jointspace trajectories (i.e., length changes of PMAs) obtained in Fig. \ref{fig:Fig7_LengthTrajectories} should be converted into actuation pressure trajectories and input to PMAs via the experimental setup (Fig. \ref{fig:Fig8_ExperimentalSetup}) to obtain the SRS locomotion. We adopted the jointspace -- pressure mapping reported in \cite{arachchige2021soft} to generate corresponding pressure inputs.
	
	\subsection{Validation of Sidewinding Gait}\label{subsec:Testing_Sidewinding}
	
	We tested the SRS's ability to achieve sidewinding locomotion by applying different pressure amplitude and frequency combinations. We initiated the testing at pressure ceiling, $p=3~bar$, and actuation frequency, $f=0.25~Hz$. We chose those values because we found that low pressures $\left(p<3~bar\right)$ onto PMAs could not provide an adequate bending in SRS sections and low frequencies $\left(f<0.25~Hz\right)$ could not provide an adequate forward momentum for moving, hence the robot was unable to show meaningful locomotion. Thus, the SRS testing was repeated by increasing the pressure ceiling and frequency by $0.25~bar$ and $0.05~Hz$ steps, respectively. We observed that the SRS can achieve a fairly stable sidewinding gait at $4~bar$ -- $1.00~Hz$. It was further revealed that high frequencies $\left(f>1.25~Hz\right)$ result in incomplete sidewinding trajectories because those exceed the operational bandwidth of PMAs. On the other hand, a high-pressure ceiling $\left(p>4~bar\right)$ overbent sections result in twisting hence the distorted trajectories. The SRS movements were video recorded using a fixed camera station. Fig. \ref{fig:Fig9_RollingSidewindingExperiments}A shows the progression of the SRS during its sidewinding movement at $4~bar-1.00~Hz$ pressure ceiling -- frequency combination. Complete videos of the experiments are given in our supplementary file. 
	
	\subsection{Validation of Helical Rolling Gait}\label{subsec:Testing_3D-Rolling}
	For testing helical rolling gait, we adopted the same procedure applied in the sidewinding testing. Accordingly, we initiated the testing at $p=3~bar$ pressure and $f=0.25~Hz$ frequency combination, and increased values on a trial-and-error basis. It must be noted that the choice for the pressure ceiling depends on the properties of custom-made PMAs and overall SRS assembly including the length of pressure supply tubes. Similar to the sidewinding testing, the SRS recorded its best helical rolling trajectory replication at $4~bar$ pressure amplitude. However, unlike the sidewinding trajectory, we observed that the SRS was capable of achieving helical rolling throughout the applied frequency range $\left(0.25~Hz<f<1~Hz\right)$ at all times. When the frequency was gradually increased from $0.25~Hz$ to $1~Hz$, the out-of-plane bending amplitude of the rolling trajectory (i.e., the displacement along the $Z$ axis of $\{O\}$) decreased. This was expected since low-frequency actuation allows PMAs to realize the desired bending profile -- i.e., at a low actuation rate, air pressure reaches to PMAs in due time through long pressure supply tubes. Fig. \ref{fig:Fig9_RollingSidewindingExperiments}B shows the progression of the SRS during its helical rolling at $4~bar-0.5~Hz$, pressure -- frequency combination in which the SRS showed the best gait replication. Please refer to the accompanying supplementary video file that shows the helical rolling gait at low-medium-high frequencies and different amplitudes.
	
	\subsection{Gait Analysis}\label{subsec:Gait_Analysis}
	
	We employed the image processing method (perspective image projection) reported in \cite{arachchige2021soft} to estimate the robot displacement on the actuated plane (on the X-Y plane) using video feedback and geometric blocks on the carpeted floor as illustrated in Fig. \ref{fig:Fig10_LocomotionTracking}. Correspondingly, we computed the SRS velocities shown in Table \ref{Table:LocomotionPerformance}.  
	It shows that the SRS replicated the sidewinding locomotion faster than the helical rolling. This is obvious since the SRS replicated sidewinding trajectory only at higher frequencies $\left(\approx~1~Hz\right)$. On the contrary, the SRS replicated its best helical rolling trajectory at mid-range frequencies $\left(\approx 0.5~Hz\right)$. 
	As expected, the SRS maintained skin-ground contacts at three points during helical rolling (Fig. \ref{fig:Fig10_LocomotionTracking}B). The SRS was expected to maintain skin-ground contacts at two points during sidewinding locomotion (Fig. \ref{fig:Fig10_LocomotionTracking}A). In reality (as witnessed in experimental videos), even though the SRS touches the ground at more points, the skin-ground contact has been dramatically reduced. 
	
	\begin{table} [tb]
		\setlength{\tabcolsep}{5pt}
		\centering
		\caption{\textsc{Locomotion Performance in Each Gait}}
		\label{Table:LocomotionPerformance}
		\begin{tabular}{|l|c|c|} 
			\hline
			\multirow{3}{*}{Gait Type} & \multicolumn{2}{c|}{Travelling velocity} \\
			& \multicolumn{2}{c|}{$[cms^{-1}]$} \\ 
			\cline{2-3}
			& $V_x$ & $V_y$ \\ 
			\hline
			Sidewinding~ & 13.38 & 14.12 \\ 
			\hline
			Helical rolling & 04.56 & 07.27 \\
			\hline
		\end{tabular}
	\end{table}

	\section{Conclusions}
	We proposed a wheelless soft robotic snake that utilizes spatial deformation to achieve snake locomotion gaits, namely sidewinding and helical rolling, without friction anisotropy. We reported the four-bending section (12 DoF) robot construction and developed a complete, floating-base kinematic model. We derived the parametric mathematical models for the aforementioned locomotion gaits and discretized the backbone curves thereof for one cycle. Those curves were then projected to the SRS coordinate frame and jointspace curves for each were derived using an optimization-based inverse kinematic approach. The joint space trajectories were then mapped to pressure trajectories and applied to the prototype SRS to experimentally validate the wheelless locomotion gaits under various parametric values (i.e., cyclic frequencies and amplitudes). The SRS deformation matched well with the mathematical models and demonstrated the feasibility and efficacy of spatial deformation in achieving wheelless SRS locomotion. Our future research will focus on dynamic modeling and validation of spatial locomotion of SRSs.

	\bibliographystyle{IEEEtran}
	\bibliography{refs}
	
\end{document}